\documentclass[10pt,twocolumn,letterpaper]{article}

\usepackage{cvpr}
\usepackage{times}
\usepackage{epsfig}
\usepackage{graphicx}
\usepackage{amsmath}
\usepackage{amssymb}

\usepackage{bbm}
\usepackage{color}
\usepackage{booktabs}
\usepackage{multirow}

\def\bx{\mathbf{x}}

\def\bv{\mathbf{v}}

\def\bq{\mathbf{q}}

\def\bh{\mathbf{h}}

\def\bw{\mathbf{w}}
\def\bW{\mathbf{W}}

\def\bM{\mathbf{M}}
\def\bm{\mathbf{m}}

\def\br{\mathbf{r}}
\def\bo{\mathbf{o}}
\def\bp{\mathbf{p}}
\def\by{\mathbf{y}}

\newcommand{\first}[1]{\textcolor{red}{#1}}
\newcommand{\second}[1]{\textcolor{blue}{#1}}
\newcommand{\third}[1]{\textcolor{cyan}{#1}}

\usepackage[pagebackref=false,breaklinks=true,letterpaper=true,colorlinks,urlcolor=magenta,citecolor=blue,linkcolor=blue,bookmarks=false]{hyperref}

\cvprfinalcopy 

\def\cvprPaperID{2089} 

\ifcvprfinal\fi
\begin{document}

\title{Visual Question Answering with Memory-Augmented Networks}

\author{Chao Ma, Chunhua Shen\thanks{C. Shen is the corresponding author.}, ~Anthony Dick, Qi Wu, Peng Wang, Anton van den Hengel, and Ian Reid \\
	Australian Institute for Machine Learning, and Australian Centre for Robotic
	Vision　\\	
	 The University of Adelaide  \\	  
	{\tt\small c.ma@adelaide.edu.au}
}
\maketitle

\begin{abstract}
	
	
	
	In this paper, we exploit memory-augmented neural networks to predict accurate answers to visual questions, even
	when those answers rarely occur in the training set. The
	memory network incorporates both internal and external
	memory blocks and selectively pays attention to each training exemplar. We show that memory-augmented neural networks are able to maintain a relatively long-term memory
	of scarce training exemplars, which is important for visual
	question answering due to the heavy-tailed distribution of
	answers in a general VQA setting. Experimental results
	in two large-scale benchmark datasets show the favorable
	performance of the proposed algorithm with the comparison to
	state of the art.
	
\end{abstract}

\section{Introduction}

Given an open-ended question and a reference image, the task of visual question answering (VQA) is to predict an answer to the question that is consistent with the image. Existing VQA systems train deep neural networks to predict answers, where  \textit{image-question} pairs are jointly embedded as training data, and answers are encoded as one-hot labels. 
Despite significant progress in recent  years~\cite{antol2015vqa,gao2015you,malinowski2015ask,ren2015image,jabri2016revisiting}, 
this approach does not scale well to completely general, freeform visual question answering. There are two main reasons for this.


\begin{figure}[t]
	\small	
	\includegraphics[width=.23\textwidth]{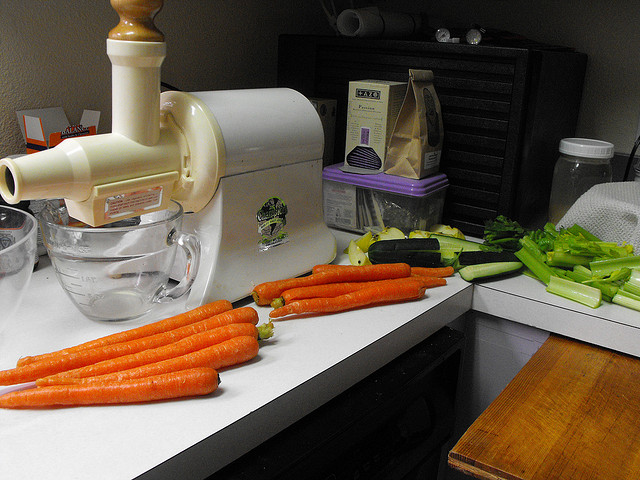} \
	\includegraphics[width=.23\textwidth]{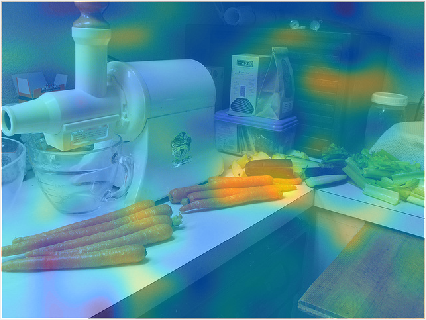} \\
	\textbf{Q}: What is the \second{dark} \third{green} \first{vegetable}? \\
	\textbf{A}: Cucumber (Ours) ~~~~ \textbf{A}: Broccoli \cite{DBLP:conf/nips/LuYBP16} ~~~~ \textbf{A}: Lettuce \cite{antol2015vqa} \\
	\caption{An example on the VQA benchmark~\cite{antol2015vqa}. Given an input question and the reference image, our method takes both visual attention and textual attention into account, and predicts a more accurate answer than recent baseline systems. We highlight the top 3 textual weights by \first{red}, \second{blue}, and \third{cyan}.}
	\label{fig:demo}
\end{figure}

First, deep models trained with gradient based methods learn to respond to the majority of training data rather than specific scarce exemplars. However, the distribution of question and answer pairs in natural language tends to be heavy tailed. By its definition, VQA involves a wide variety of question and answer topics that cannot be predicted in advance. In fact, the words that human observers are interested in are often unknown or rare \cite{Gulcehre2016unkownwords}. Figure \ref{fig:demo} shows one example, in which baseline VQA systems \cite{antol2015vqa,DBLP:conf/nips/LuYBP16} exclude the rare answer \textit{cucumber} from training set and therefore fail to predict a reasonable answer for the test question \textit{What it the dark green vegetable?}. Despite the rare words being very important to human observers, when evaluating error on a per-answer basis, excluding the rare words from training sets altogether can often improve the overall performance. Because of this, existing approaches mark rare words in questions as meaningless unknown tokens (e.g., \textit{unk}) and simply exclude rare answers from training set.   

Second, existing VQA systems learn about the properties of objects from question-answer pairs, sometimes independently of the image. Taking the question \textit{How many zebras are in the image?} as an example, VQA algorithms are subject to bias in human language without the underlying models truly understanding the visual content. For instance, the number of zebras,  which has appeared in training answers, provides a strong prior in predicting answers.
%
While it would be desirable to learn to count from images, this remains an open problem. In the meantime, being able to better exploit words and concepts from the heavy tails of the textual question and answer distributions would enable more accurate question answering involving these less common words. 
It is therefore of great importance to selectively pay more attention to the heavy-tailed answers during the training stage as shown in Figure \ref{fig:fre}.

We take inspiration from the recent development of memory-augmented neural networks \cite{DBLP:conf/icml/SantoroBBWL16} as well as the co-attention mechanism \cite{DBLP:conf/nips/LuYBP16} between image and question pairs. Neural networks with augmented external memory \cite{DBLP:conf/icml/SantoroBBWL16,kaiser2017learning} are able to reason over extremely scarce training data, while attention mechanisms have become dominantly popular in natural language processing related tasks such as machine translation \cite{DBLP:journals/corr/BahdanauCB14}, image captioning \cite{xu2015show}, and visual question answering \cite{DBLP:conf/nips/LuYBP16}. In this work, we align our motivation to the recent work \cite{kaiser2017learning} using memory networks to remember rare events and propose to learn memory-augmented networks with attention to rare answers for VQA.
We first employ a co-attention mechanism to jointly embed image and question features together. We then learn memory-augmented networks that maintain a long-term memory of scarce training data to facilitate VQA. 
Note that our method significantly differs from the dynamic memory network \cite{xiong2016} in that our memory networks contain both an internal memory inside LSTM and an external memory controlled by LSTM, while Xiong et al. \cite{xiong2016} only implement the memory inside an attention gated recurrent unit (GRU). We summarize the main contributions of this work as follows: 

\begin{figure}
	\centering
	\includegraphics[width=.4\textwidth]{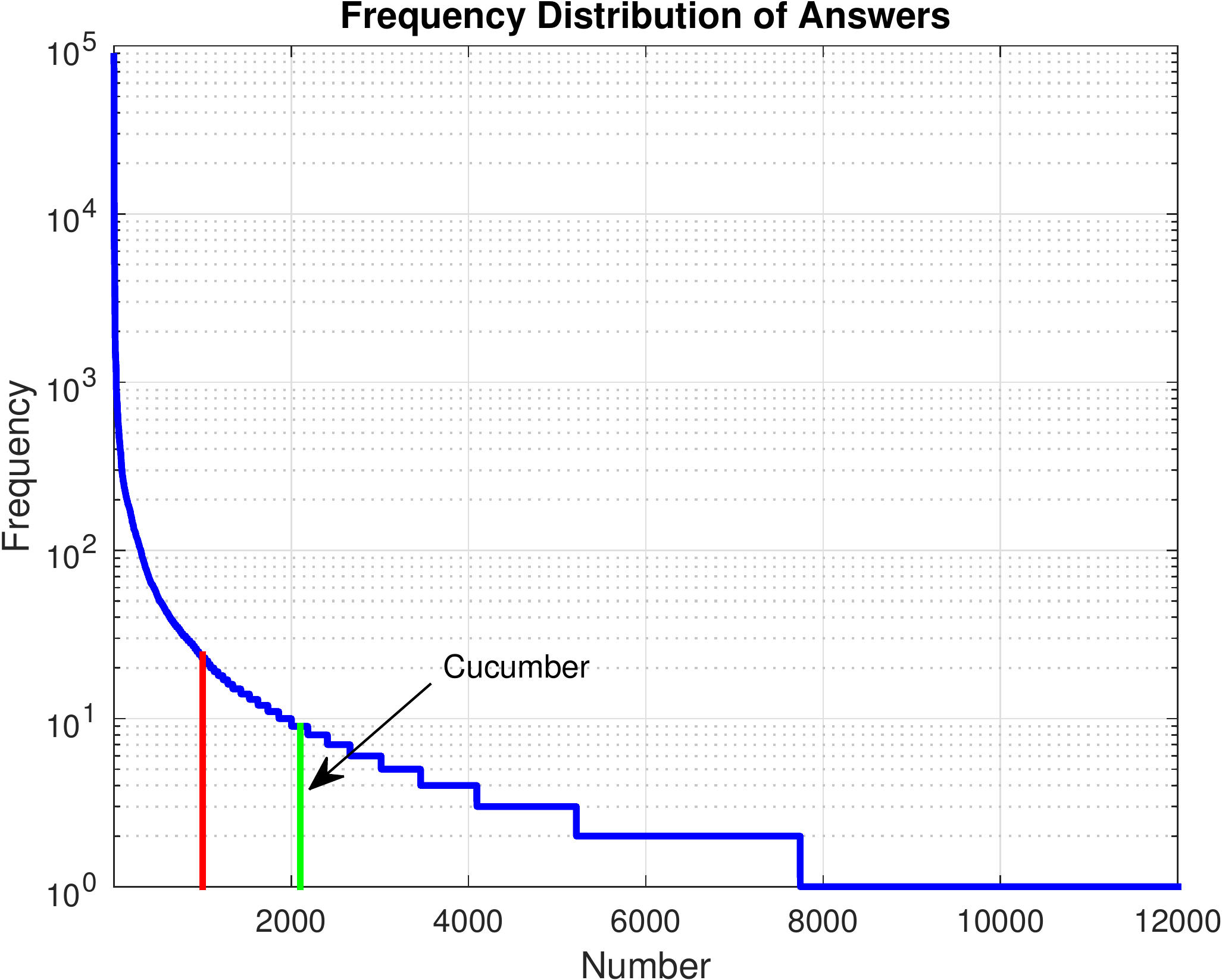}
	\caption{Frequency distribution of answers on the VQA dataset~\cite{antol2015vqa}. Existing VQA systems~\cite{antol2015vqa,DBLP:conf/nips/LuYBP16} typically select top 1000 answers (red line) as one-hot labels to train deep networks. A large number of valuable  answer and question pairs, such as the answer \textit{Cucumber} in Figure \ref{fig:demo}, are excluded. }
	\label{fig:fre}
\end{figure}

\begin{itemize}
	\item We propose to use memory-augmented networks to increase our capacity to remember uncommon question and answer pairs for visual question answering. 
	\item We use a co-attention mechanism to attend to the most relevant image regions as well as textual words before jointly embedding the image and question features. 
	\item We validate the proposed algorithm on two benchmark datasets. Experimental results show that the proposed algorithm performs favorably against state of the art. 
\end{itemize}

\section{Related Work}
{\flushleft \bf Joint Embedding.} 
Existing approaches mainly cast VQA as a multi-label classification problem. Since the VQA task is to answer a question regarding a reference image, VQA requires multi-model reasoning over visual and textual data. A large number of recent approaches \cite{antol2015vqa,gao2015you,malinowski2015ask,ren2015image,jabri2016revisiting} explore a joint embedding to represent image and question pairs using deep neural networks. Typically, image features are the outputs of the last fully connected layer of convolutional neural networks (CNNs) that are pre-trained on object recognition datasets. A textual question is split into sequential words, which are fed into a recurrent neural network (RNN) to yield a fixed-length feature vector as question representation. The image and question features are jointly embedded as one vector to train multi-label classifiers that predict answers. Numerous efforts have been made to improve the effectiveness of joint embedding, such as exploring external knowledge~\cite{wu2015ask}, multi-model compact pooling~\cite{fukui2016multimodal}, multimodel residual learning~\cite{DBLP:conf/nips/KimLKHKHZ16}, or fixed language embedding using bag of words \cite{zhou2015simple}. Note that these approaches generate features over entire images and questions, thus cannot attend to the most relevant regions and textual words to facilitate VQA.

\begin{figure*}[t!]
	\centering
	\includegraphics[width=.92\textwidth]{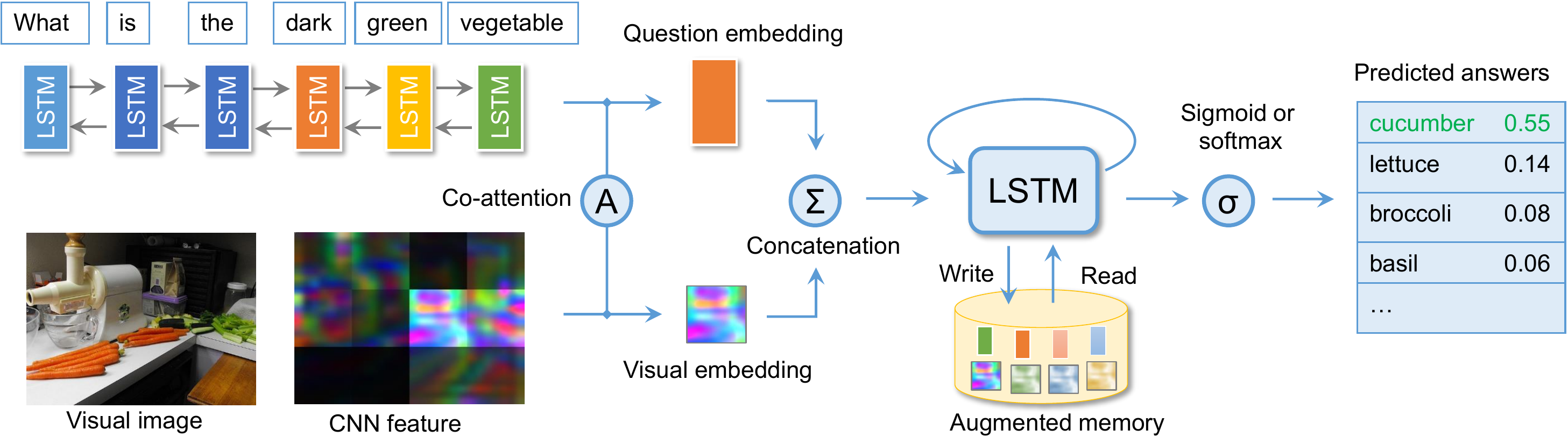}
	\vspace{1mm}
	\caption{Flowchart of the proposed algorithm. We use the last pooling layer of pre-trained CNNs to extract image features that encode spatial layout information. We employ bi-directional LSTMs to generate a fixed-length feature vector for each word. A co-attention mechanism attends to relevant image regions and textual words. We concatenate the attended image and question feature vectors and feed them into a memory-augmented network, which consists of a standard LSTM as controller and an augmented external memory. The controller LSTM determines when to write or read from the external memory. The memory-augmented network plays a key role in maintaining a long-term memory of scarce training data. We take the outputs of the memory-augmented network as final embedding for the image and question pair, and feed this embedding into a classifier to predicts answers.}
	\label{fig:overview}
\end{figure*}

{\flushleft \bf Attention Mechanism.} 
Instead of directly using the entire-image embedding from the fully
connected layer of a deep CNN, attention models have been widely used to select the most relevant image regions for VQA.  The attention mechanism~\cite{andreas2015deep} typically consists of a language parser and a number of CNN feature vectors representing spatially distributed regions. The language parser determines the relevance of each region to answer the question. Yang et al. \cite{yang2015stacked} perform image attention multiple times in a stacked manner to infer the answer progressively. In \cite{xu2015ask}, the authors use a multi-hop image attention mechanism to capture fine-grained information from the question. \cite{DBLP:conf/cvpr/ShihSH16} applies off-the-shelf region proposal algorithms to generate object regions, and select the most relevant regions to predict answers. In~\cite{xiong2016}, the authors propose an attention based GRU to facilitate answer retrieval. In addition to visual attention, the recent work~\cite{DBLP:conf/nips/LuYBP16} proposes a co-attention mechanism with question attention as well. Similarly, we apply a co-attention mechanism attending to both image regions and textual words in questions. But unlike~\cite{DBLP:conf/nips/LuYBP16}, that treats each word in a sentence independently, we take into account the sequential consistency of textual words, i.e., we first use a bi-directional LSTMs to generate word embeddings, and perform the question attention scheme on the output embeddings.

{\flushleft \bf Memory Network.}  
Since Weston et al.~\cite{weston2014} proposed a
memory component over simple facts for the question answering problem, memory networks have become increasingly popular in language processing. Memory networks generally consist of input, scoring, attention and response components.  In~\cite{Sukhbaatar2015},  Sukhbaatar et al. train memory networks in an end-to-end manner, which does not require labeling supporting facts during the training stage, unlike earlier networks~\cite{weston2014}. In \cite{kumar2015}, Kumar et al. build memory networks on neural sequence models with attention. Given a question, a neural attention mechanism allows memory networks to selectively pay attention to specific inputs.     
This benefits a wide range of computer vision and language processing problems, such as image classification~\cite{DBLP:conf/nips/StollengaMGS14}, image caption~\cite{xu2015show} and machine translation~\cite{cho2014learning,DBLP:journals/corr/BahdanauCB14,DBLP:conf/emnlp/LuongPM15}. 
Other recent neural architectures with memory or attention include neural Turing machines~\cite{DBLP:journals/corr/GravesWD14}, stack-augmented RNNs~\cite{DBLP:conf/nips/JoulinM15}, and hierarchical memory networks~\cite{Chandar2016}. In view of the great potential of memory networks for VQA~\cite{DBLP:conf/icml/SantoroBBWL16,xiong2016}, we propose to use a memory-augmented neural network to selectively pay more attention to heavy-tailed question and answer pairs. For implementation, we use LSTM to control reading from and writing to an augmented external memory. Our memory networks thus significantly differs from the attention GRU network in \cite{xiong2016}.

\section{Proposed Algorithm}

We show the main steps of the algorithm in Figure \ref{fig:overview}. Given an input question and reference image, we use a co-attention mechanism to select the most relevant image regions and textual words in questions. Specifically, we use the outputs of the last pooling layer of pre-trained CNNs (VGGNet \cite{simonyan2014vgg} or ResNet \cite{DBLP:conf/cvpr/HeZRS16}) as image features, which maintain spatial layout information. We split the question into sequential words, which are fed into bi-directional LSTMs to generate sequentially fixed-length word embeddings. The co-attention mechanism computes weights for each CNN feature vector as well as each textural word embedding (see Figure \ref{fig:overview} the highlighted weights in different colors). We concatenate the relevant image and question features as an embedding of image and question pair. We use the standard LSTM network as a controller, which determines when to read from and write to an external memory. Note that our memory networks essentially contains two memory blocks: an internal memory inside LSTM and an external memory controlled by LSTM. The memory-augmented networks maintain a long-term memory of heavy-tailed question answers. We use the outputs of the memory-augmented networks for training classifiers that predict answers.

In the rest of this section, we first introduce the image feature extractor using pre-trained CNNs, as well as the question encoder using bi-directional LSTMs. Following that, we present the sequential co-attention mechanism that attends to the most relevant image regions and textual words. We then present the used memory-augmented network in more details. We discuss the answer reasoning scheme at the end of this section. 

\subsection{Input Representation}

{\flushleft \bf Image Embedding.} We use the pre-trained VGGNet-16 \cite{simonyan2014vgg} and ResNet-101 \cite{DBLP:conf/cvpr/HeZRS16} to extract CNN features. Following \cite{DBLP:conf/nips/LuYBP16,yang2015stacked}, we resize images to
$448\times448$ before feeding them into CNNs. We take the outputs of the last pooling layer of VGGNet-16 (\textit{pool5}) or the layer under the last pooling layer of ResNet-101 (\textit{res5c}) as image features corresponding to $14\times 14$ spatially distributed regions. We denote the output features by $\{\bv_1,\ldots,\bv_N\}$, where $N=196$ is the total number of regions and $\bv_n$ is the $n$-th feature vector with the dimension of 512 for VGGNet-16 or 2048 for ResNet-101.

{\flushleft \bf Question Embedding.} As shown in Figure \ref{fig:overview}, we exploit bidirectional LSTMs \cite{DBLP:journals/tsp/SchusterP97} to generate question features. Let $\{\bw_1,\ldots, \bw_T\}$ be the one-hot vector of input question with $T$ words. We apply an embedding matrix $\bM$ to embed
the words to another vector space as $\bx_t = \bM\bw_t$. The embedded vectors are then fed into bidirectional LSTMs:
\begin{eqnarray}
\bh_t^+&=&\text{LSTM}(\bx_t,\bh_{t-1}^+) \\
\bh_t^-&=&\text{LSTM}(\bx_t,\bh_{t+1}^-)
\end{eqnarray}
where $\bh_t^+$ and $\bh_t^-$ denote the hidden states of the forward and backward LSTMs at time $t$, respectively. We concatenate the bi-directional states $\bh_t^+$ and $\bh_t^-$  as feature vector $\bq_t=[\bh_t^+,\bh_t^-]$ to represent the $t$-th word in input question. 

\subsection{Sequential Co-Attention}
\label{sec:coatt}
\begin{figure}
	\centering
	\includegraphics[width=.42\textwidth]{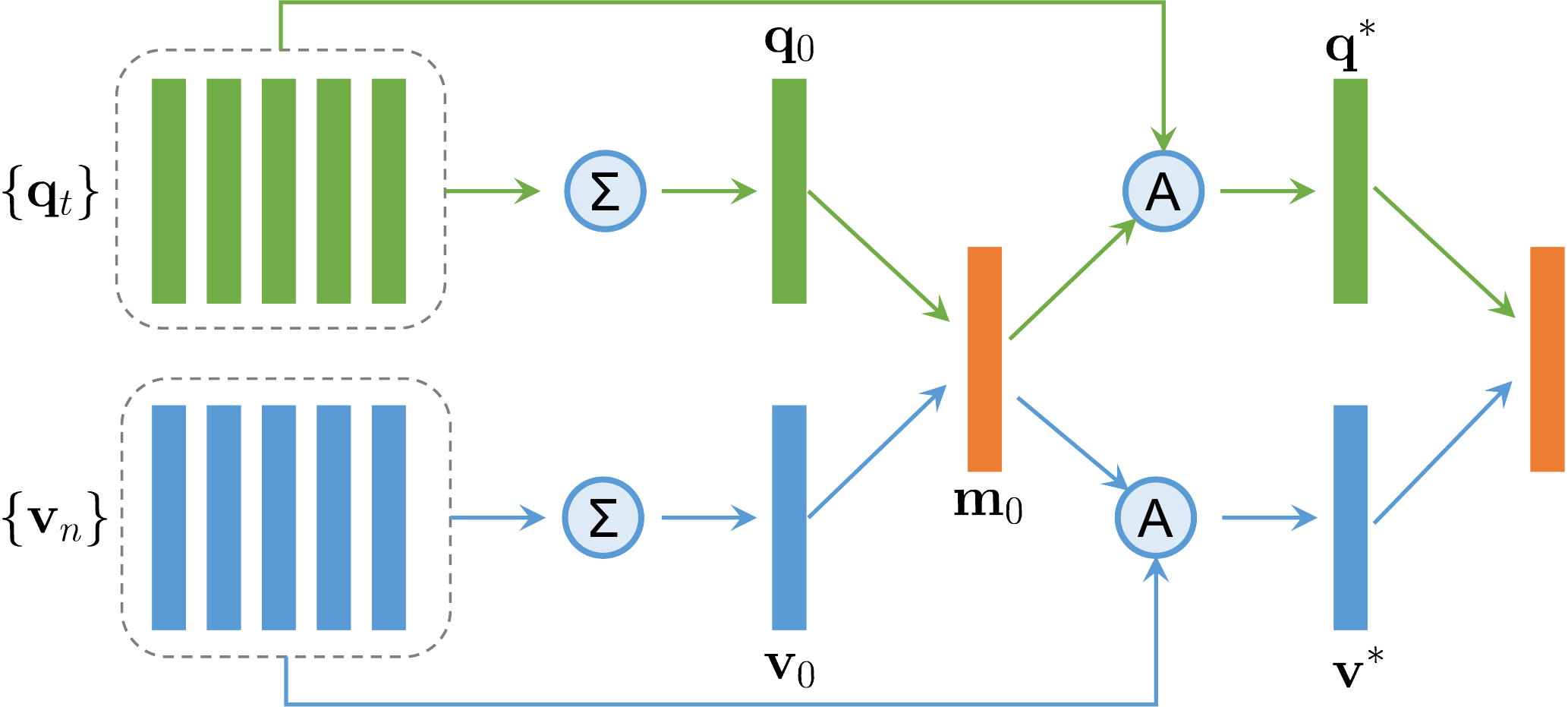}
	\vspace{1mm}
	\caption{Sequential co-attention mechanism.}
	\label{fig:att}
\end{figure}

Given a pair of visual and question feature vectors, the co-attention mechanism aims to attend to the most relevant parts of each type of features referring to the other.    
Let $\{\bv_n\}$ and $\{\bq_t\}$ be the visual and question feature vectors respectively. We compute a base vector $\bm_0$ to advise the later attention computation as follows:
\begin{eqnarray}
\bm_0&=&\bv_0 \odot \bq_0 \\ 
\text{where} \quad \bv_0&=& \tanh(\frac{1}{N}\sum_{n}\bv_n) \\
\bq_0&=& \frac{1}{T}\sum_{t}\bq_t
\end{eqnarray}
Here $\odot$ is the element-wise product. To ensure that the visual feature vector $\bv_0$ and question feature vector $\bq_0$ have the same dimension, we always set the size of hidden states of bidirectional LSTMs to one half of the dimension of the visual feature vector $\bv_n$. We conduct the co-attention mechanism identically on the visual and question feature vectors. We implement the soft attention using a two-layer neural network. For visual attention, the soft weights  $\{\alpha_n|n=1,\ldots,N\}$  and the attended visual feature vector $\bv^*$ are as follows:   
\begin{eqnarray}
\bh_n &=&\tanh(\bW_\bv\bv_n)\odot\tanh(\bW_\bm\bm_0) \\
\alpha_n &=& \text{softmax}(\bW_\bh\bh_n)  \\
\bv^*&=&\tanh(\sum_{n=1}^{N}\alpha_n\bv_n)
\end{eqnarray}
where $\bW_\bv$,$\bW_\bm$ and $\bW_\bh$ denote hidden states. Similarly, we compute the attended question feature vector $\bq^*$ as follows:
\begin{eqnarray}
\bh_t &=&\tanh(\bW_\bq\bq_t)\odot\tanh(\bW_\bm\bm_0) \\
\alpha_t &=& \text{softmax}(\bW_\bh\bh_t)  \\
\bq^*&=&\sum_{t=1}^{T}\alpha_t\bq_t
\end{eqnarray}
We concatenate the attended vectors $\bv^*$ and $\bq^*$ to represent the input image and question pair. Figure \ref{fig:att} illustrates the pipeline of the co-attention mechanism. 

\subsection{Memory Augmented Network}

Once we obtain the feature vector by concatenating the visual and question features, instead of directly applying this vector to learning classifier as in~\cite{antol2015vqa,gao2015you,malinowski2015ask,ren2015image,jabri2016revisiting}, we would like a mechanism to determine the importance of this exemplar in terms of position and training ordering among the whole training data. In other words, we hope such a mechanism can selectively pay more attention to scarce training items whose effect is always neglected during a huge amount of training iterations. Xiong et al.  \cite{xiong2016} have shown that RNNs can function as an attention mechanism to select the most relevant parts of the current input referring to the hidden state from previous time-steps. However, the RNNs lack external memory to maintain a long-term memory for scarce training data. 

In this work, we propose to use a memory-augmented neural network \cite{DBLP:conf/icml/SantoroBBWL16} for VQA. Specifically, we exploit the standard LSTM network to serve as a controller that receives input data and interacts with an external memory module using a number of read and write heads. The LSTM has internal memory and functions as similarly as the attention GRU in \cite{xiong2016}. But our memory-augmented network significantly differs from \cite{xiong2016} in that, in addition to the internal memory in LSTM, we exploit an external memory, $\bM_t$, that is both read from and written to. Specifically, the memory block $\bM_t$ consists of a set of row vectors as memory slots. 

Let $\{\bx_t,\by_t\}$, $t=1,\ldots,T$, be the overall $T$ training data, where $\bx_t=[\bv_t^*,\bq_t^*]$ denotes the concatenation of visual and question feature vectors and $\by_t$ is the corresponding one-hot encoded answer vector. We first feed the feature vector $\bx_t$ into the LSTM controller as:
\begin{equation}
\label{eq:lstm-vq}
\bh_t=\text{LSTM}(\bx_t,\bh_{t-1})
\end{equation} 
For reading from external memory $\bM_t$, we take the hidden state $\bh_t$ as a query for $\bM_t$. First, we compute the cosine distance for the query vector $\bh_t$ and each individual row in memory: 

\begin{equation}
D(\bh_t,\bM_t(i))=\frac{\bh_t\cdot\bM_t(i)}{\|\bh_t\|\|\bM_t(i)\|}
\end{equation}
Next, we compute a read weight vector $\bw_t^r$ using a softmax over the cosine distance: 
\begin{equation}
\label{eq:wr}
w_t^r(i)=\text{softmax}\big(D(\bh_t,\bM_t(i)\big) 
\end{equation}
With these read-weights, a new retrieved memory $\br_t$ is obtained as follows:
\begin{equation}
\br_t=\sum_{i}w_t^r(i)\bM_i
\label{eq:read}
\end{equation}
Finally, we concatenate the new memory vector $\br_t$ with the controller hidden state $\bh_t$ together to produce the output vector $\bo_t$ for learning classifier.

Previous memory networks \cite{xiong2016,Sukhbaatar2015,peng2015languageunderstanding} rely on a learnable gate to write the new input data in specific memory addresses. Despite the favorable performance in sequence-based prediction tasks, this scheme is not optimal for the VQA task that deals with a conjunctive coding of heavy-tailed training data rather than a sequence. In this case, we would like the writer to strike a balance between writing new information to rarely used location and writing to recently used location. Similarly to \cite{DBLP:conf/icml/SantoroBBWL16}, we employ the usage weights $\bw_t^u$ to control writing to memory. We update the usage weights $\bw_t^u$ at each time-step by decaying its previous state and adding the current read and write weights:
\begin{equation}
\bw_t^u=\gamma\bw_{t-1}^u+\bw_t^r+\bw_t^w.
\label{eq:decaying}
\end{equation}
Here, $\gamma$ is a decay parameter and $\bw_t^r$ is computed in \eqref{eq:wr}.
To compute the write weights, we introduce a truncation scheme to update the least-used positions. Note that this truncation scheme facilitates networks to maintain a longer term of memory for heavy-tailed training data rather than to rapidly erase their effect. Here we use the notation $m(\bv, n)$ to denote the $n$-th smallest element of a vector $\bv$. We apply a learnable sigmoid gate parameter to compute a convex combination of the previous read weights and usage weights:
\begin{equation}
\bw_t^w=\sigma(\alpha)\bw_{t-1}^r+\big(1-\sigma(\alpha)\big) \mathbbm{1}\big(\bw_{t-1}^u\le m(\bw_{t-1}^u,n)\big)  
\label{eq:write}
\end{equation}
Here, $\sigma(\cdot)$ is a sigmoid function, $\frac{1}{1+e^{-x}}$, and $\alpha$ is a scalar gate parameter. The indicator function $\mathbbm{1}(x)$ returns 1 if $x$ is true otherwise 0. 
A larger $n$ results in maintaining a longer term of memory of scarce training data. Compared to the internal memory inside LSTM, both the parameters $\gamma$ and $n$ can adjust the rate of writing to external memory. This gives us more freedom to regulate memory update. The output hidden state $\bh_t$ in \eqref{eq:lstm-vq} is written to memory in accordance with the write weights:
\begin{equation}
\bM_t^i= \bM_{t-1}(i)+w_t^w(i)\bh_t
\end{equation}

\subsection{Answer Reasoning}

We concatenate the hidden state $\bh_t$ in \eqref{eq:lstm-vq} and the reading memory $\br_t$ in \eqref{eq:read} as the final embedding $\bo_t=[\bh_t,\br_t]$ for the image and question pair. We pass $\bo_t$ to generate output distribution. Specifically, we use a one-layer perceptron that consists of a linear hidden layer as well as a softmax function to output categorical distribution. The categorical distribution yields a vector $\bp_t$ whose elements show the class probabilities:
\begin{eqnarray}
\bh_t&=&\tanh(\bW_\bo\bo_t)\\
\bp_t&=&\text{softmax}(\bW_\bh\bh_t)
\end{eqnarray}
Here $\bW_\bo$ and $\bW_\bh$ are the hidden parameters of the linear layer. In the training stage, given the output distribution $\bp_t$, the network is optimized by minimizing the loss over the input one-hot encoded label vector $\by_t$:
\begin{equation}
\mathcal{L}(\theta)=-\sum_{t}\by_t^\top\log\bp_t
\end{equation}

\begin{table*}
	\small
	\centering
	\caption{Single model accuracy on the VQA benchmark \cite{antol2015vqa}. The proposed algorithm performs favorably against state-of-the-art methods in both the multiple-choice and open-ended tasks. The best and second best values are highlighted by bold and underline.}
	\vspace{1mm}
	\label{tb:vqa}
	\resizebox{\textwidth}{!}{
	\setlength{\tabcolsep}{5pt}
	\begin{tabular}{lcccccccccccccccc}
		\toprule
		& \multicolumn{8}{c}{Test-dev}                                         & \multicolumn{8}{c}{Test-standard}                                    \\ \cmidrule(lr){2-9} \cmidrule(lr){10-17}
		& \multicolumn{4}{c}{Multiple-choice} & \multicolumn{4}{c}{Open-ended} & \multicolumn{4}{c}{Multiple-choice} & \multicolumn{4}{c}{Open-ended} \\ \cmidrule(lr){2-5}\cmidrule(lr){6-9}\cmidrule(lr){10-13}\cmidrule(lr){14-17}
		Method                                      & Y/N    & Num   & Other & All   & Y/N     & Num     & Other  & All    & Y/N    & Num   & Other & All   & Y/N     & Num     & Other  & All    \\ \midrule
		iBOWIMG \cite{zhou2015simple}                  & 76.7            & 37.1 & 54.4  & 61.7 & 76.6       & 35.0 & 42.6  & 55.7 & 76.9            & 37.3 & 54.6  & 62.0 & 76.8       & 35.0 & 42.6  & 55.9 \\
		DPPnet \cite{Noh2015image}                     & 80.8            & 38.9 & 52.2  & 62.5 & 80.7       & 37.2 & 41.7  & 57.2 & 80.4            & 38.8 & 52.8  & 62.7 & 80.3       & 36.9 & 42.2  & 57.4 \\
		VQA team \cite{antol2015vqa}                   & 80.5            & 38.2 & 53.0  & 62.7 & 80.5       & 36.8 & 43.1  & 57.8 & 80.6            & 37.7 & 53.6  & 63.1 & 80.6       & 36.4 & 43.7  & 58.2 \\
		SAN \cite{yang2015stacked}                     & -               & -    & -     & -    & 79.3       & 36.6 & 46.1  & 58.7 & -               & -    & -     & -    & -          & -    & -     & 58.9 \\
		NMN \cite{andreas2015deep}                     & -               & -    & -     & -    & 80.5       & 37.4 & 43.1  & 57.9 & -               & -    & -     & -    & -          & -    & -     & 58.0 \\
		ACK \cite{wu2015ask}                           & -               & -    & -     & -    & 81.0       & 38.4 & 45.2  & 59.2 & -               & -    & -     & -    & 81.1       & 37.1 & 45.8  & 59.4 \\
		SMem \cite{xu2015ask}                     & -               & -    & -     & -    & 80.9       & 37.3 & 43.1  & 58.0 & -               & -    & -     & -    & 80.8       & 37.3 & 43.1  & 58.2 \\
		DMN+ \cite{xiong2016}                     & -               & -    & -     & -    & 80.5       & 36.8 & 48.3  & 60.3 & -               & -    & -     & -    & -          & -    & -     & 60.4 \\
		MRN-ResNet \cite{DBLP:conf/nips/KimLKHKHZ16}       & \textbf{82.4}            & 39.7 & 57.2  & 65.6 & \underline{82.4}       & 38.4 & 49.3  & 61.5 & \textbf{82.4}            & 39.6 & 58.4  & 66.3 & \textbf{82.4}       & \textbf{38.2} & 49.4  & 61.8 \\
		Re-Ask-ResNet \cite{malinowski2015ask}    & -               & -    & -     & -    & 78.4       & 36.4 & 46.3  & 58.4 & -               & -    & -     & -    & 78.2       & 36.3 & 46.3  & 58.4 \\
		HieCoAtt-ResNet \cite{DBLP:conf/nips/LuYBP16} & 79.7            & 40.0 & 59.8  & 65.8 & 79.7       & 38.7 & 51.7  & 61.8 & -               & -    & -     & 66.1 & -          & -    & -     & 62.1 \\
		RAU-ResNet \cite{noh2016training}         & \underline{81.9}            & \underline{41.1} & 61.5  & 67.7 & 81.9       & \textbf{39.0} & 53.0  & 63.3 & \underline{81.7}            & \underline{40.0} & 61.0  & 67.3 & \underline{81.7}       & \textbf{38.2} & 52.8  & 63.2 \\
		MCB-ResNet \cite{fukui2016multimodal}     & -               & -    & -     & \underline{69.1} & \textbf{82.5}       & 37.6 & \textbf{55.6}  &\textbf{ 64.7} & -               & -    & -     & -    & -          & -    & -     & -    \\
		MLP-ResNet \cite{jabri2016revisiting} & -               & -    & -     & -    & -       & - & -  & - & 80.8 & 17.6 & 62.0 & 65.2   & -          & -    & -     & - \\
		VQA-Mac-ResNet \cite{wang2017vqamachine}  & 81.5            & 40.0 & 62.2  & 67.7 & 81.5       & 38.4 & 53.0  & 63.1 & 81.4            & 39.8 & \underline{62.3}  & \underline{67.8} & 81.4       & \textbf{38.2} & \underline{53.2}  & \underline{63.3} \\\midrule
		Ours-VGG                                     & 81.1            & 41.0 & \underline{62.5}  & 67.8 & 81.2       & 37.8 & 50.7  & 61.8 & 81.2            & 39.3 & 61.7  & 67.4 & 81.2       & 36.4 & 51.7  & 62.3 \\
		Ours-ResNet                                    & 81.6            & \textbf{42.1} & \textbf{65.2}  & \textbf{69.5} & 81.5       & \textbf{39.0} & \underline{54.0}  & \underline{63.8} & 81.6            & \textbf{40.9} & \textbf{65.1}  & \textbf{69.4} & \underline{81.7}       & 37.6 & \textbf{54.7}  & \textbf{64.1} \\ \bottomrule
	\end{tabular}
}
\end{table*}

\section{Experiments}

\subsection{Implementation Details}

We fix all the parameters throughout experimental validations. The dimension of every hidden layer including word embedding, LSTMs, and attention models is set to 512. 
The learning rates for question embedding and answering modules are set to $3\times10^{-3}$ and $3\times10^{-4}$, respectively. The decaying parameter $\gamma$ in \eqref{eq:decaying} is $10^{-4}$. The truncation number $n$ in \eqref{eq:write} is set to 4. We train our networks using the Adam optimization scheme \cite{DBLP:conf/iclr/KingmaB14}. We decay the learning rates every epoch by a factor of 0.9. We add gradient noises from Gaussian distribution to improve network learning as in \cite{DBLP:journals/corr/NeelakantanVLSK15}. To deal with exploding gradients, we perform gradient clipping by limiting the gradient magnitude to 0.1. CNN feature extractor builds upon VGGNet-16 \cite{simonyan2014vgg} and ResNet-101  \cite{DBLP:conf/cvpr/HeZRS16} without fine-tuning. We implement our networks using Torch7~\cite{Collobert_torch7}.  

\subsection{VQA Benchmark}

The VQA benchmark dataset~\cite{antol2015vqa} contains approximately 200K real images from the MSCOCO dataset~\cite{lin2014microsoft}. Each image corresponds to three questions, and each question has ten answers collected from human subjects. The dataset is typically divided into four splits: 248,349 training questions, 121,512
validation questions, 60,864 developing test questions, and 244,302 standard test questions. We note that, in view of the issue of data imbalance, the VQA benchmark has been recently updated with more balanced data~\cite{balanced_vqa_v2}. As this work deals with the rare question answer pairs, we mainly evaluate our method using the previous VQA benchmark~\cite{antol2015vqa}. We train our model using the training and validation splits, and report the test results on both the developing (\textit{test-dev}) and standard (\textit{test-standard}) splits. The VQA benchmark includes two different tasks: multiple-choice and open-ended, which do or do not provide a set of candidate answers, respectively. 
 For both tasks, we follow~\cite{antol2015vqa} to report the accuracy as: 
\begin{equation}
\text{Acc}(\hat{a})=\min\bigg\{\frac{\#\text{humans that labeled }\hat{a}}{3},1\bigg\}
\end{equation}
where $\hat{a}$ is the predicted answer.

{\flushleft \bf Overall Performance.} 
We compare our method with state-of-the-art VQA algorithms in Table \ref{tb:vqa}. For fairness, we compare with single model approaches on the VQA benchmark leader-board, and highlight the methods using the ResNet~\cite{DBLP:conf/cvpr/HeZRS16} for image features. The remaining methods all use the VGGNet \cite{simonyan2014vgg} except the iBOWING method~\cite{zhou2015simple}, which uses the GoogLeNet \cite{szegedy2014googlenet} to extract image features.  
The proposed method performs favorably against state of the art in both the open-ended and multiple-choice tasks. 
Among the state-of-the-art methods, the MCB method \cite{fukui2016multimodal} using a multi-model compact pooling scheme with attention mechanism performs best in the \textit{test-dev} validation, while our algorithm achieves a higher overall accuracy for the multiple-choice task. It is worth mentioning that even with VGGNet features, our method generally performs well against the state-of-the-art algorithms with ResNet features. The HieCoAtt method \cite{DBLP:conf/nips/LuYBP16} exploits a similar co-attention mechanism to emphasize the most relevant image regions and textural words as we do. However, it does not incorporate an augmented external memory and thus does not make full use of scarce training exemplars. The DMN+ method \cite{xiong2016} builds a dynamic memory network based on RNNs, but it only exploits the internal memory inside RNNs rather than an augmented external memory. With the use of memory-augmented networks and sequential co-attention mechanism, our method reaches substantially higher accuracy than the HieCoAtt and DMN+ methods in all test settings: The performance gains in the \textit{test-dev} validation are 3.7\% (multiple-choice)  and 2.0\% (open-ended) compared to HieCoAtt \cite{DBLP:conf/nips/LuYBP16}, and 3.5\% (open-ended) compared to DMN+ \cite{xiong2016}. For the \textit{test-standard} validation, our method advances state-of-the-art accuracy achieved by the recently proposed VQA machine \cite{wang2017vqamachine} from 67.8\% to 69.4\% for the multiple-choice task, and from 63.3\% to 64.1\% for the open-ended task respectively.

\begin{table}
	\small
	\caption{Ablation studies on the VQA benchmark \cite{antol2015vqa} using the \textit{test-dev} validation. For baseline algorithms with VGGNet \cite{simonyan2014vgg} and ResNet \cite{DBLP:conf/cvpr/HeZRS16} features, we use top 1000, 2000 and 3000 answers to train neural networks. EM indicates whether an external memory is enabled. The use of augmented external memory generally improves accuracy along with including an increasing number of answers. The best and second best values are highlighted by bold and underline.}
	\vspace{1mm}
	\label{tb:ablation}
	\resizebox{0.48\textwidth}{!}{
	\setlength{\tabcolsep}{3pt}
	\begin{tabular}{cccccccccc}
		\toprule
		\multicolumn{2}{c}{}                                                        & \multicolumn{4}{c}{Multiple-choice} & \multicolumn{4}{c}{Open-ended} \\ \cmidrule(lr){3-6}\cmidrule(lr){7-10}
		& EM & Y/N     & Num    & Other   & All    & Y/N   & Num   & Other  & All   \\ \midrule
		\multirow{2}{*}{\begin{tabular}[c]{@{}c@{}}VGGNet\\ 1000\end{tabular}}    & Y  & \underline{81.4}    & 39.2   & 58.9    & 66.0   & \underline{81.3}  & 37.0  & 49.6   & 61.3  \\
		& N  & 81.0    & 37.2   & 58.5    & 65.4   & 81.0  & 34.5  & 48.6   & 60.4  \\ \midrule
		\multirow{2}{*}{\begin{tabular}[c]{@{}c@{}}VGGNet\\ 2000\end{tabular}}    & Y  & 81.3    & 40.9   & 61.6    & 67.5   & 81.2  & 37.8  & 50.7   & 61.8  \\
		& N  & 81.1    & 38.7   & 61.0    & 66.9   & 81.1  & 35.6  & 49.2   & 60.8  \\ \midrule
		\multirow{2}{*}{\begin{tabular}[c]{@{}c@{}}VGGNet\\ 3000\end{tabular}}    & Y  & 81.1    & 41.0   & 62.5    & 67.8   & 81.1  & 37.6  & 50.6   & 61.7  \\
		& N  & 80.7    & 39.7   & 62.1    & 67.3   & 80.6  & 35.5  & 49.5   & 60.8  \\ \midrule
		\multirow{2}{*}{\begin{tabular}[c]{@{}c@{}}ResNet\\ 1000\end{tabular}} & Y  & 81.0    & \underline{41.3}   & 64.0    & 68.5   & 80.9  & \underline{38.6}  & \underline{53.6}   & 63.2  \\
		& N  & 81.0    & \underline{41.3}   & 63.7    & 68.4   & 80.9  & \underline{38.6}  & 52.8   & 62.8  \\ \midrule
		\multirow{2}{*}{\begin{tabular}[c]{@{}c@{}}ResNet\\ 2000\end{tabular}} & Y  & 81.2    & 40.9   & 64.2    & \underline{68.6}   & 81.1  & 37.9  & 53.8   & \underline{63.3}  \\
		& N  & 81.2    & 40.9   & 63.3    & 68.2   & 81.2  & 37.9  & 52.2   & 62.5  \\ \midrule
		\multirow{2}{*}{\begin{tabular}[c]{@{}c@{}}ResNet\\ 3000\end{tabular}} & Y  & \textbf{81.6}   & \textbf{42.1}   & \textbf{65.2}    & \textbf{69.5}   & \textbf{81.5}  & \textbf{39.0}  & \textbf{54.0}   & \textbf{63.8}  \\
		& N  & 80.9    & 40.0   & \underline{64.6}    & \underline{68.6}   & 80.8  & 36.2  & 52.8   & 62.5  \\ \bottomrule
	\end{tabular}
}
\end{table}

{\flushleft \bf Ablation Studies.}
To evaluate the effectiveness of memory-augmented networks in dealing with heavy-tailed answers, we conduct ablation studies on the \textit{test-dev} validation. We train neural networks with and without the augmented external memory over VGGNet and ResNet features respectively, using the top 1000, 2000, and 3000 answers. Table \ref{tb:ablation} shows that the networks without an augmented external memory do not favor including more answers. Taking the baseline using ResNet features as an example, the performance using top 2000 answers drops when compared to that using top 1000 answers. This clearly shows that including more heavy-tailed answers negatively affects the overall accuracy of deep neural networks without augmented memory. 
With the use of an augmented external memory, including more heavy-tailed answers generally improves the overall accuracy in both the multi-choice and open-ended tasks. All baseline algorithms are substantially improved by enabling the augmented external memory. Interestingly, the performance gains become larger with deeper CNN features and with more answers in both the multiple-choice and open-end tasks. Fox example, for the baseline algorithm with ResNet features and top 1000 answers, enabling the augmented external memory only increases the overall accuracy by less than 0.4\%, but it increases the overall accuracy by about 1\% for the baseline algorithm using ResNet features and top 3000 answers. 

It is worth mentioning that, in the VQA literature, element-wise product, element-wise addition, outer product, and concatenation are four widely used schemes to integrate image and question embeddings. We empirically find element-wise product performs well for constructing base vectors in our scheme. Table 2 shows that our method even without external memory (ResNet-1000-N) still achieves performance gains of 2.6\% and 1\% for the multiple-choice and open-ended tasks when compared to \cite{fukui2016multimodal}.

\begin{figure}
	\centering
	\includegraphics[width=.22\textwidth]{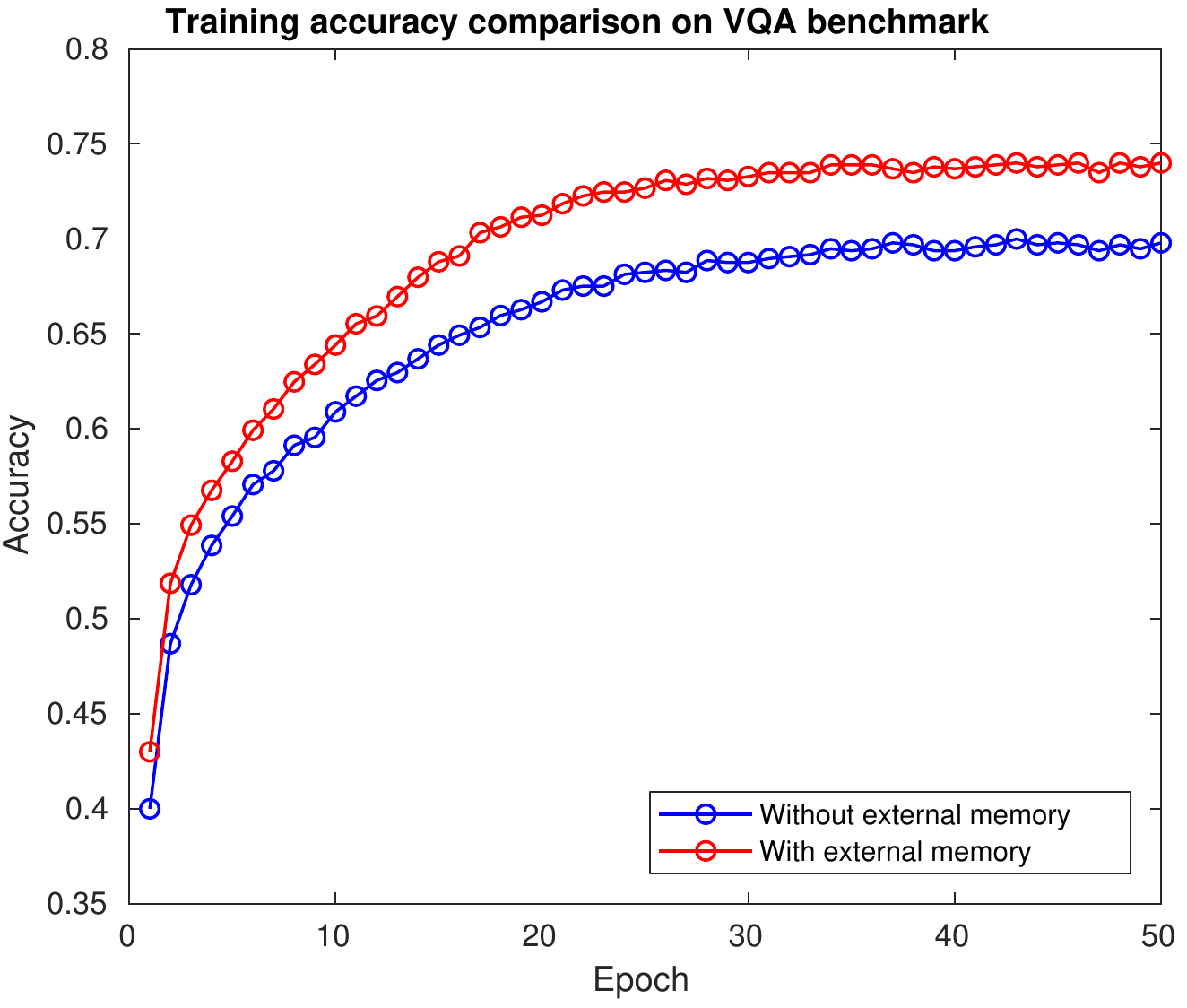}  \includegraphics[width=.22\textwidth]{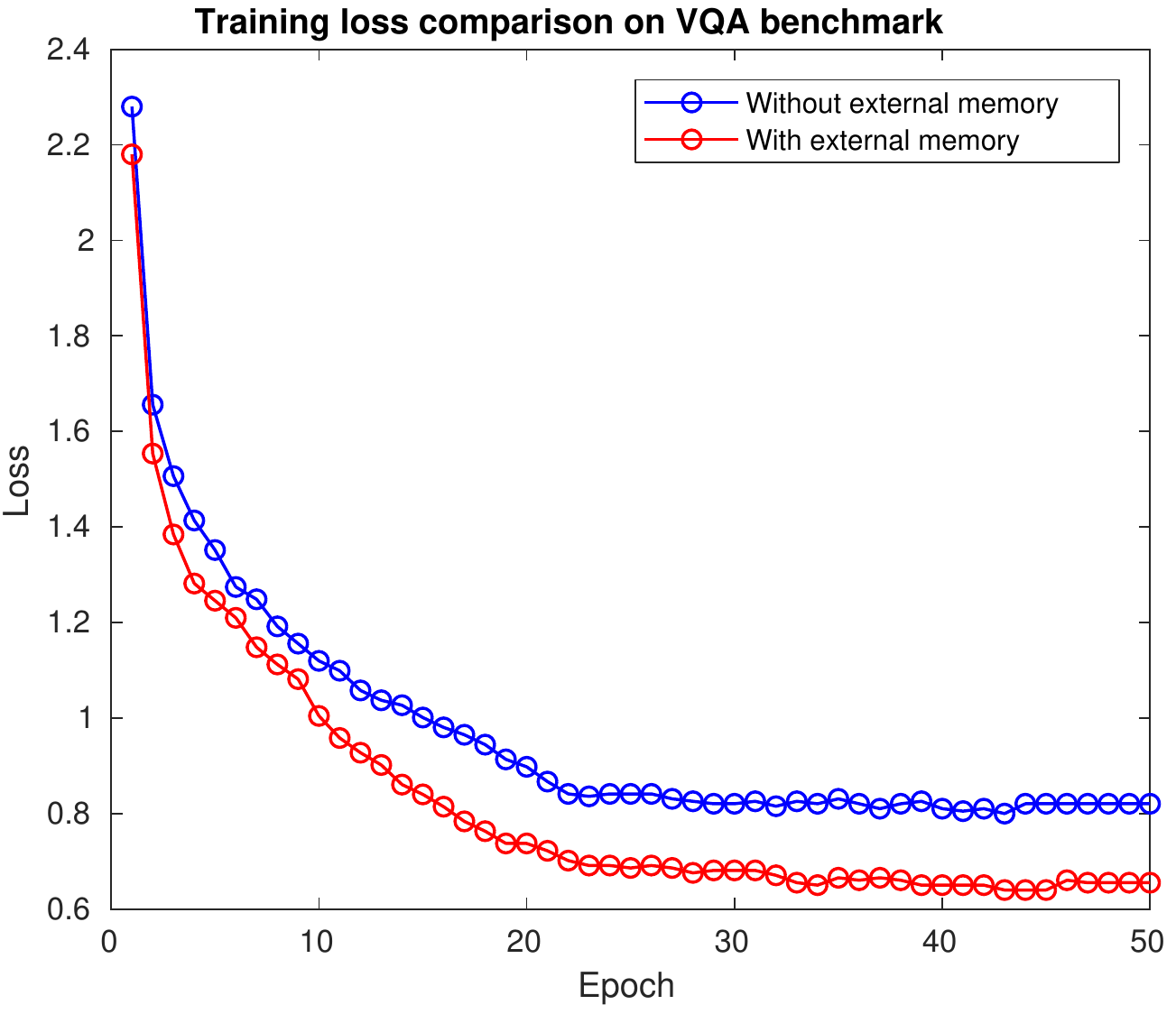}
	\caption{Accuracy and loss with or without external memory using CNN (VGGNet) image embedding with top 3000 answers for training on the \textit{train}+\textit{val} sets of the VQA benchmark.}
	\label{fig:acc-loss}
\end{figure}

To validate our design choices, we start with the CNN+LSTM baseline, which uses LSTM and CNN (VGG) as question and image embeddings as in \cite{antol2015vqa}. We incrementally add more components to the baseline model and report the performance improvement in Table \ref{tb-component}. Figure \ref{fig:acc-loss} compares the training accuracy and loss using the external memory or not.

\begin{table}
	\centering
	\small
	\setlength{\tabcolsep}{10pt}
	\caption{Component analysis. We report the overall results of multiple-choice in the \textit{test-dev} validation). We start with the CNN+LSTM baseline, which uses VGGNet image embedding and top 3000 answers for learning LSTM. We incrementally add components to obtain better performance.}
	\label{tb-component}
	\begin{tabular}{cc}
		\toprule
		Component & Accuray (\%) \\\midrule
		Forward LSTM (h+) & 63.6 \\
		+ Backwward LSTM (h-) & 64.3 \\
		+ Co-attention & 65.9 \\
		+ External memory & 67.8 \\ \bottomrule
		
	\end{tabular}
\end{table}

\vspace{-1mm}
{\flushleft \bf Answer Distribution.} Table \ref{tb:vqadata} shows the heavy-tailed distribution of the question-answer pairs on the VQA benchmark dataset \cite{antol2015vqa}. Among the 369,861 (248,349 from the training split and 121,512 from the validation split) question-image pairs, including more question-answer pairs to train deep neural networks is usually more challenging due to the noisy training data. We list the representative accuracy of the multiple-choice task in the \textit{test-dev} validation.

\begin{table}
	\centering
	\small
	\caption{Heavy-tailed distribution in the training and validation sets of the VQA benchmark dataset \cite{antol2015vqa}. $x$-axis shows the number of selected answers for training deep network. Including more question/answer pairs makes optimizing deep models more challenging. We exploit memory-augmented networks to make full use of the heavy-tailed answers to facilitate visual question answering.}
	\label{tb:vqadata}
	\setlength{\tabcolsep}{6pt}
	\begin{tabular}{ccccc} \toprule
		& 1000  & 2000  & 3000  & 4000  \\ \midrule
		Number & 320,029 & 334,554 & 341,814 & 346,287 \\
		Ratio  & 0.865 & 0.905 & 0.924 & 0.936 \\
		Accuracy & 0.685 & 0.686 & 0.695 & 0.693 \\
		\bottomrule
	\end{tabular}
\end{table}

\begin{figure}
	\small
	\includegraphics[width=.22\textwidth]{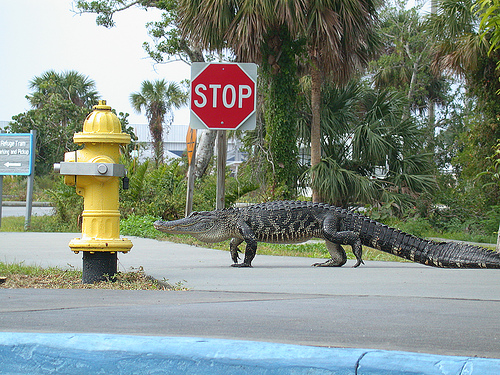}~
	\includegraphics[width=.22\textwidth]{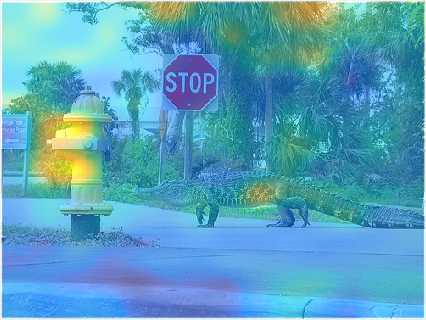} \\
	\textbf{Q}: What is the \first{yellow} \second{object} on the \third{left}? \\
	\textbf{A}: Fire hydrant (Ours) ~~~ \textbf{A}: Stop sign \cite{DBLP:conf/nips/LuYBP16} ~~~ \textbf{A}: Stop sign \cite{antol2015vqa} \\	
	\includegraphics[width=.22\textwidth]{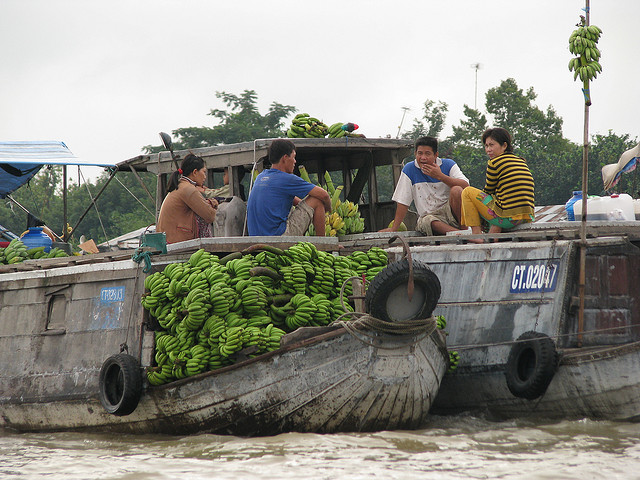}~
	\includegraphics[width=.22\textwidth]{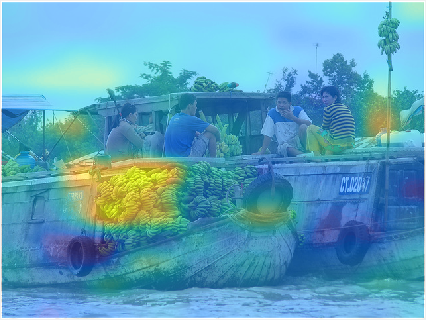} \\
	\textbf{Q}: \third{What} \first{fruit} is \second{showing} in this picture?  \\
	\textbf{A}: Bananas (Ours) ~~~~~ \textbf{A}: Bananas \cite{DBLP:conf/nips/LuYBP16} ~~~~~ \textbf{A}: Bananas \cite{antol2015vqa} \\
	\includegraphics[width=.22\textwidth]{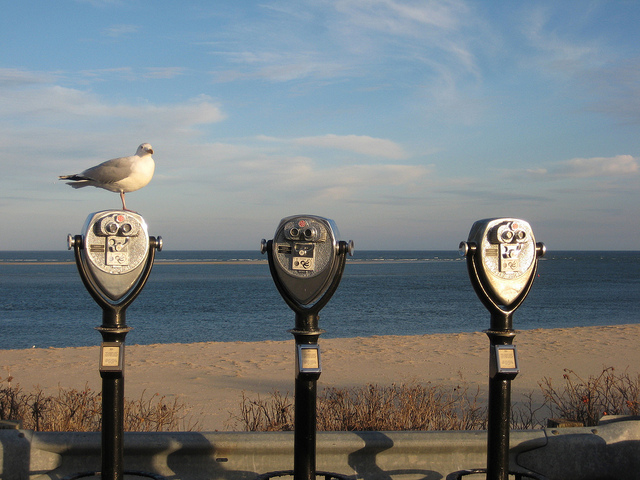}~
	\includegraphics[width=.22\textwidth]{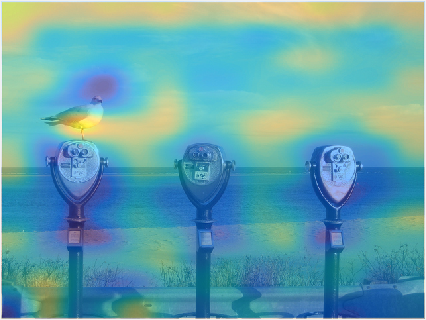}\\
	\textbf{Q}: \third{Where} is the \first{bird} \second{standing} on?  \\
	\textbf{A}: Sand (Ours) ~~~~ \textbf{A}: Parking meter \cite{DBLP:conf/nips/LuYBP16} ~~~~ \textbf{A}: Road \cite{antol2015vqa} \\
	
	\caption{Qualitative results on the VQA benchmark dataset \cite{antol2015vqa}. 
		We visualize the image attention map and highlight the first 3 relevant words in each question by \first{red}, \second{blue} and \third{cyan}. Our method selectively pays attention to important regions and words that are critical to answer the questions correctly. 
	}
	\label{fig:mscoco}
\end{figure}

\vspace{-1mm}
{\flushleft \bf Qualitative Evaluation.} Figure \ref{fig:mscoco} shows the qualitative results. We visualize the image attention as well as the textual word attention computed by the sequential co-attention mechanism (see Section \ref{sec:coatt}). The proposed method selectively pays attention to important regions and words and predicts correct answers to challenging questions that require fine-grained reasoning. The last example in Figure \ref{fig:mscoco} shows a failure case, where a bird is standing on a telescope. To the question \textit{What is the bird standing on?}, our method fails to attend to the proper region of the telescope and provides an unexpected answer \textit{sand}.

\vspace{-1mm}
{\flushleft \bf VQA v2.0.} We mainly use VQA v1.0~\cite{antol2015vqa} to evaluate our method as it contains heavily imbalanced questions/answers, which can highlight the effectiveness of our method in exploiting imbalanced training data. For completeness, we report the results on VQA v2~\cite{balanced_vqa_v2} in Table \ref{table1}, which shows that our method performs well against state of the art as well.  

\begin{table}[h]
	\centering
	\small
	\vspace{-2mm}
	\caption{Results on the VQA benchmark v2.0~\cite{balanced_vqa_v2}  (\emph{test-std}).}
	\label{table1}
	\begin{tabular}{ccccc}
		\toprule			
		&Y/N &  Num & Other & All \\\midrule
		MCB \cite{fukui2016multimodal} &78.8 & 38.3 & 53.3 & 62.3 \\
		Ours (ResNet-3000) & 79.2 & 39.5 & 52.6 &  62.1 \\
		\bottomrule
	\end{tabular}
\end{table} 

\vspace{-2mm}
\subsection{Visual 7W Telling}

The Visual 7W Telling dataset~\cite{DBLP:conf/cvpr/ZhuGBF16} includes $69,817$ training questions, $28,020$ validation questions, and $42,031$ test questions. Each question has four answer choices. Following \cite{DBLP:conf/cvpr/ZhuGBF16}, we report the performance by measuring the percentage of correctly answered questions. For fair comparison, we select top 5000 answers to train networks. We validate our algorithm with comparison to the state-of-the-art algorithms including LSTM+CNN \cite{antol2015vqa}, Visual 7W baseline~\cite{DBLP:conf/cvpr/ZhuGBF16}, MCB~\cite{fukui2016multimodal}, and MLP~\cite{jabri2016revisiting}. Table \ref{tb:v7w} shows the results of this comparison. Note that the top performing MLP \cite{jabri2016revisiting} and MCB \cite{fukui2016multimodal} algorithms both use the word2vec scheme \cite{DBLP:conf/nips/MikolovSCCD13} to generate fixed question embedding. Jabri et al. \cite{jabri2016revisiting} mention that it is more helpful to use a fixed question embedding on the Visual 7W dataset \cite{DBLP:conf/cvpr/ZhuGBF16} as its size is relatively small when compared to the VQA benchmark~\cite{antol2015vqa}. We first learn our memory-augmented network as well as the question embedding from scratch. We report the overall accuracy as 59.4\%. Our method still performs well against the Visual 7W baseline \cite{DBLP:conf/cvpr/ZhuGBF16}, which affirms the advantage of memory-augmented network for VQA, as the Visual 7W baseline \cite{DBLP:conf/cvpr/ZhuGBF16} does not incorporate external memory. In addition, we borrow the training and validation questions from the VQA dataset \cite{antol2015vqa} for training networks. With the use of larger training data, our method performs well compared to the MCB method~\cite{fukui2016multimodal}, whereas we learn the question embedding from scratch.

\begin{table}[t]
	\centering
	\small
	\caption{Accuracy on the Visual 7W Telling \cite{DBLP:conf/cvpr/ZhuGBF16} dataset. We train the question embedding from scratch. Our method peforms favorably against state-of-the-art methods. The best and second best values are highlighted by bold and underline.}
	\label{tb:v7w}
	\resizebox{0.48\textwidth}{!}{
		\setlength{\tabcolsep}{2pt}
		\begin{tabular}{lccccccc} \toprule
			& What   & Where  & When   & Who    & Why    & How    & \multicolumn{1}{c}{\multirow{2}{*}{Overall}} \\ 
			Methods                        & 47.8\% & 16.5\% & 4.5\%  & 10.0\% & 6.3\%  & 14.9\% & \multicolumn{1}{c}{}                         \\ \midrule
			LSTM+CNN \cite{antol2015vqa} & 48.9   & 54.4   & 71.3   & 58.1   & 51.3   & 50.3   & 52.1                                         \\
			Visual 7W \cite{DBLP:conf/cvpr/ZhuGBF16}                   & 51.5   & 57.0   & 75.0   & 59.5   & 55.5   & 49.8   & 55.6                                         \\
			MCB \cite{fukui2016multimodal}                           & 60.3   & \underline{70.4}   & \underline{79.5}   & \underline{69.2 }  & \underline{58.2}   & 51.1   & 62.2                                         \\
			MLP \cite{jabri2016revisiting}                           & \textbf{64.5}   & \textbf{75.9}   & \textbf{82.1}   & \textbf{72.9}   & \textbf{68.0}   & \textbf{56.4}   & \textbf{67.1}                                         \\ \midrule
			Ours   & 59.0       &  63.2       &    75.7    &   60.3    &   56.2     &   52.0     &   59.4                                           \\
			Ours + VQA              & \underline{62.2} & 68.9 & 76.8 & 66.4 & 57.8 & \underline{52.9} & \underline{62.8}  \\ \bottomrule                                           
		\end{tabular}
	}
	\vspace{-1mm}
\end{table}

\vspace{-1mm}
\section{Conclusion}
We make a first attempt to explicitly address the issue of rare concepts in visual question answering. 
The main pipeline of the proposed algorithm consists of a co-attention module to select the relevant image regions and textual word features, as well as a memory module that selectively pays attention to rare training data. 
An LSTM module plays a role of controller who determines when to write or read from the external memory block. The outputs of the augmented LSTM are the features for learning a classifier that predicts answers. The proposed algorithm performs well against state-of-the-art VQA systems on two large-scale benchmark datasets, and is demonstrated to successfully answer questions involving rare concepts where other VQA methods fail.

\vspace{-3mm}
\paragraph{Acknowledgments.} We gratefully acknowledge the support of the Australian Research Council through the Centre of Excellence for Robotic Vision CE140100016 and  Laureate Fellowship FL130100102 to I. Reid.

{\small
	\bibliographystyle{ieee}
	\bibliography{vqabib}
}

\end{document}